\documentclass[11pt,a4paper]{article}
\usepackage[hyperref]{acl2021}
\usepackage{times}
\usepackage{latexsym}

\usepackage{epsfig}
\usepackage{graphicx}
\usepackage{amsmath}
\usepackage{amssymb}
\usepackage{multirow}
 \usepackage{bm}
 \usepackage{amsmath}
 \usepackage{soul}
 \setstcolor{red}
 \usepackage{pbox}
\usepackage{graphicx, array, blindtext}

 \newcommand{\name}{{\texttt{MedWriter}}} 
 \usepackage{amsthm}
 
\usepackage{microtype}

\aclfinalcopy 

\title{Writing by Memorizing: Hierarchical Retrieval-based Medical Report Generation}

\author{\small Xingyi Yang \thanks{\;\;This work was done when Xingyi Yang remotely worked with Fenglong Ma.}\\
\small UC San Diego\\
\small \texttt{x3yang@eng.ucsd.edu}\\
\And 
\small Muchao Ye\\
\small Penn State University\\
\small \texttt{muchao@psu.edu}\\
\And
\small Quanzeng You\\
\small Microsoft Azure Computer Vision\\
\small \texttt{quyou@microsoft.com}\\
\And 
\small Fenglong Ma\thanks{\;\;Corresponding author}\\
\small Penn State University\\
\small \texttt{fenglong@psu.edu}}

\date{}

\begin{document}
\maketitle
\begin{abstract}
Medical report generation is one of the most challenging tasks in medical image analysis. 
Although existing approaches have achieved promising results, they either require a predefined template database in order to retrieve sentences or ignore the hierarchical nature of medical report generation. 
To address these issues, we propose~{\name} that incorporates a novel hierarchical retrieval mechanism to automatically extract both report and sentence-level templates for clinically accurate report generation. 
{\name} first employs the Visual-Language Retrieval~(VLR) module to retrieve the most relevant reports for the given images. To guarantee the logical coherence between sentences, the Language-Language Retrieval~(LLR) module is introduced to retrieve relevant sentences based on the previous generated description.
At last, a language decoder fuses image features and features from retrieved reports and sentences to generate meaningful medical reports. We verified the effectiveness of our model by automatic evaluation and human evaluation on two datasets, i.e., Open-I and MIMIC-CXR.
\end{abstract}

\section{Introduction}
Medical report generation is the task of generating reports based on medical images, such as radiology and pathology images.
Given that this task is time-consuming and cumbersome, researchers endeavor to relieve the burden of physicians by automatically generating the findings and descriptions from medical images with machine learning techniques.

\begin{figure}[t]
    \center
   \includegraphics[width=\linewidth]{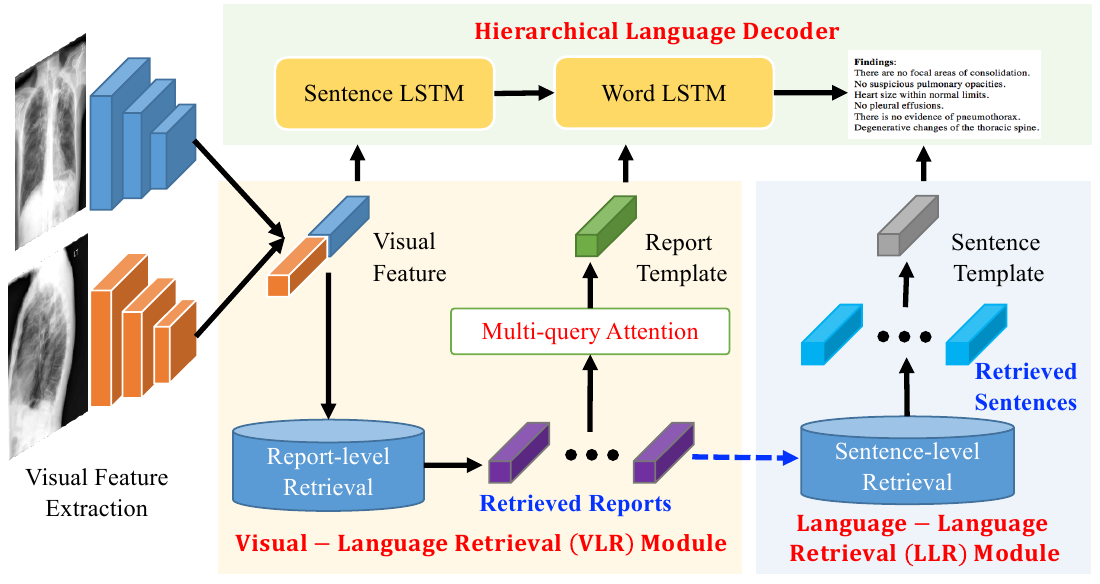}
   \vspace{-0.3in}
    \caption{Overview of the proposed {\name}.}
    \label{fig:overview}
    \vspace{-0.2in}
\end{figure}


Existing studies can be roughly divided into two categories, i.e., generation-based and retrieval-based approaches. Generation-based methods, including LRCN~\cite{donahue2015long}, CoAtt~\cite{jing2017automatic}, and MvH+AttL~\cite{yuan2019automatic}, focus on generating image captions with a encoder-decoder model that leverage image features. However, they are unable to produce linguistically diverse descriptions and depict rare but prominent medical findings. On the other hand, Retrieval-based methods such as HRGR-Agent~\cite{li2018hybrid} and KEPP~\cite{li2019knowledge}, 
pay attention to memorizing templates to generate standardized reports from \textit{a predefined retrieval database}. 
However, the quality of generated reports significantly depends on the manually curated template database.
Besides, they only use sentence-level templates for the generation but ignore to learn the report-level templates, which prevent them from generating more accurate reports.

To address the aforementioned issues, we propose a new framework called {\name} as shown in \figurename~\ref{fig:overview}. 
{\name} introduces a novel hierarchical retrieval mechanism working with a hierarchical language decoder to \textbf{automatically learn the dynamic report and sentence templates from the data} for generating accurate and professional medical reports.
{\name} is inspired by the process of how physicians write medical reports in real life. They keep report templates in mind and then generate reports for new images by using the key information that they find in the medical images to update the templates sentence by sentence.  

In particular, we use three modules to mimic this process. First, {\name} generates \textbf{report-level templates} from the Visual-Language Retrieval (VLR) module using the visual features as the queries. To generate accurate reports, {\name} also predicts disease labels based on the visual features and extracts medical keywords from the retrieved reports. We propose a \textbf{multi-query attention} mechanism to learn the report-level template representations. Second, to make the generated reports more coherent and fluent, we propose a Language-Language Retrieval (LLR) module, which aims to learn \textbf{sentence-level templates} for the next sentence generation by analyzing between-sentence correlation in the retrieved reports. 
Finally, a \textbf{hierarchical language decoder} is adopted to generate the full report using visual features, report-level and sentence-level template representations.
The designed two-level retrieval mechanism for memorization is helpful in generating accurate and diverse medical reports. To sum up, our contributions are:
\begin{itemize}
    \vspace{-0.08in}
	\item To the best of our knowledge, we are the first to model the memory retrieval mechanism in both report and sentence levels. By imitating the standardized medical report generation in real life, our memory retrieval mechanism effectively utilizes existing templates in the two-layer hierarchy in medical texts. This design allows {\name} to generate more clinically accurate and standardized reports.
	\vspace{-0.1in}
	\item On top of the retrieval modules, we design a new multi-query attention mechanism to fuse the retrieved information for medical report generation. The fused information can be well incorporated with the existing image and report-level information, which can improve the quality of generated report .
	\vspace{-0.1in}
	\item Experiments conducted on two large-scale medical report generation datasets, i.e., Open-i and MIMIC-CXR show that {\name} achieves better performance compared with state-of-the-art baselines measured by CIDEr, ROUGE-L, and BLEUs. Besides, case studies show that {\name} provides more accurate and natural descriptions for medical images through domain expert evaluation.
\end{itemize}

\section{Related work}
\paragraph{Generation-based report generation}
Visual captioning is the process of generating a textual description given an image or a video. The dominant neural network architecture of the captioning task is based on the encoder-decoder framework~\cite{bahdanau2014neural,vinyals2015show,mao2014deep}, with attention mechanism~\cite{xu2015show,you2016image,lu2017knowing,anderson2018bottom,wang2019hierarchical}. As a sub-task in the medical domain, early studies directly apply state-of-the-art encoder-decoder models as CNN-RNN~\cite{vinyals2015show}, LRCN~\cite{donahue2015long} and AdaAtt~\cite{lu2017knowing} to medical report generation task. 
To further improve long text generation with domain-specific knowledge, later generation-based methods introduce hierarchical 
LSTM with co-attention~\cite{jing2017automatic} or use the medical concept features~\cite{yuan2019automatic} to attentively guide the report generation. On the other hand, the concept of reinforcement learning~\cite{liu2019clinically} is utilized to ensure the generated radiology reports correctly describe the clinical findings.

To avoid generating clinically non-informative reports, external domain knowledge like knowledge graphs~\cite{DBLP:conf/aaai/ZhangWXYYX20,li2019knowledge} and anchor words~\cite{biswal2020clinical} are utilized to promote the medical values of diagnostic reports. CLARA~\cite{biswal2020clinical} also provides an interactive solution that integrates the doctors' judgment into the generation process.

\paragraph{Retrieval-based report generation}
Retrieval-based approaches are usually hybridized with generation-based ones to improve the readability of generated medical reports. For example, KERP~\cite{li2019knowledge} uses abnormality graphs to retrieve most related sentence templates during the generation. HRGR-Agent\cite{li2018hybrid} incorporates retrieved sentences in a reinforcement learning framework for medical report generation. 
However, they all require a template database as the model input. Different from these models, {\name} is able to automatically learn both report-level and sentence-level templates from the data, which significantly enhances the model applicability.

\begin{figure*}[!htb]
\center
    \includegraphics[width=0.9\linewidth]{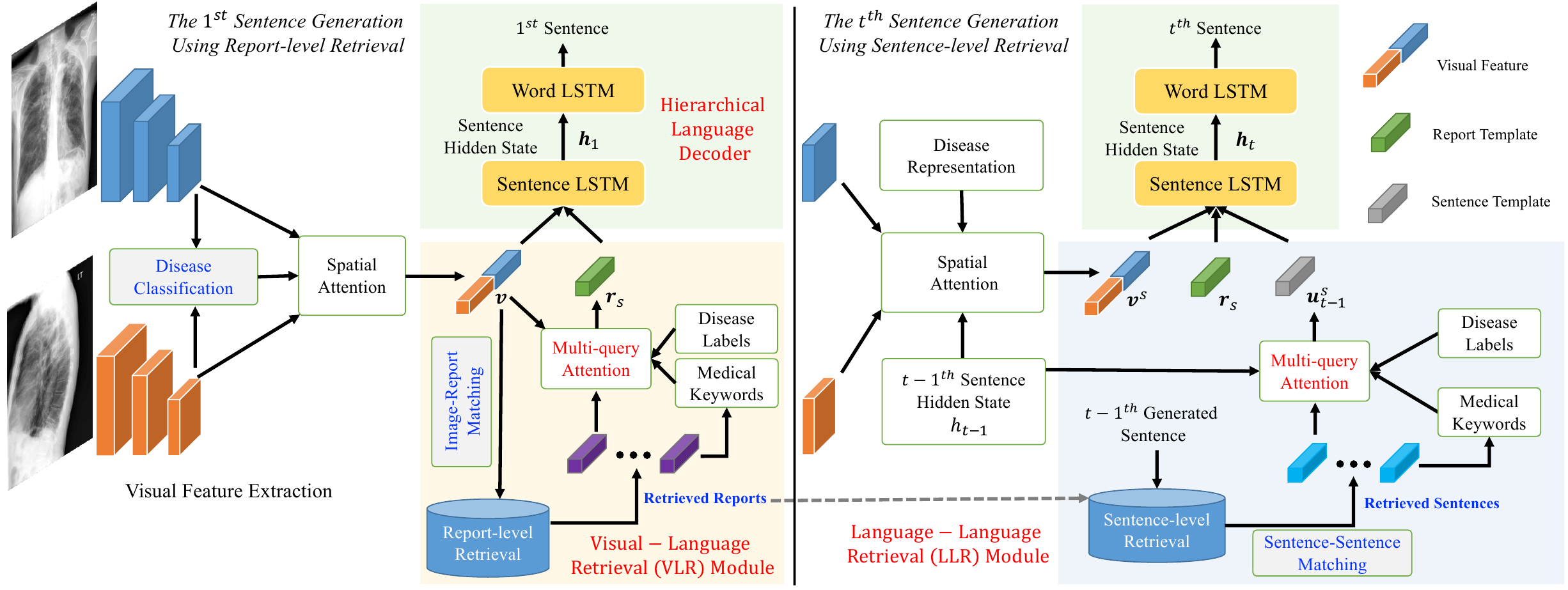}
    \vspace{-0.1in}
    \caption{Details of the proposed {\name} model for medical report generation. The left part is used to learn report template representations via the visual-language retrieval (VLR) module, which is further used to generate the first sentence via the hierarchical language decoder. The right part shows the details of the language-language (LLR) module used for generating the remaining sentences.}
    \label{fig:framework}
    \vspace{-0.1in}
\end{figure*}

\section{Method}

As shown in \figurename~\ref{fig:framework}, we propose a new framework called {\name}, which consists of three modules. 
The \textbf{Visual-Language Retrieval~(VLR)} module works on the \textit{report level} and uses visual features to find the most relevant template reports based on a multi-view image query. 
The \textbf{Language-Language Retrieval~(LLR)} module works on the \textit{sentence level} and retrieves a series of candidates that are most likely to be the next sentence from the retrieval pool given the generated language context. Finally, {\name} generates accurate, diverse, and disease-specified medical reports by a \textbf{hierarchical language decoder} that fuses the visual, linguistics and pathological information obtained by VLR and LLR modules. To improve the effectiveness and efficiency of retrieval, we first pretrain VLR and LLR modules to build up a retrieval pool for medical report generation as follows. 



\subsection{VLR module pretraining}\label{vlrm pretraining}

The VLR module aims to retrieve the most relevant medical reports from the training report corpus for the given medical images. 
The retrieved reports are further used to learn an abstract template for generating new high-quality reports. 
Towards this goal, we introduce a self-supervised pretraining task by judging whether an image-report pair come from the same subject, i.e., \textbf{image-report matching}. It is based on an intuitive assumption that an image-report pair from the same subject shares certain common semantics. 
More importantly, the disease types associated with images and the report should be similar. 
Thus, in the pretraining task, we also take disease categories into consideration.

\subsubsection{Disease classification} \label{disease classifition}
The input of the VLR module is a series of multi-modal images and the corresponding report $(\{\bm{I}_i\}_{i=1}^b,r)$ where the set $\{\bm{I}_i\}_{i=1}^b$ consists of $b$ images, and $r$ denotes the report. We employ a Convolutional Neural Network (CNN) $f_v(\cdot)$ as the image encoder to obtain the feature of a given image $\bm{I}_i$, i.e., $\bm{v}_i = f_v(\bm{I}_i)$,
where $\bm{v}_i\in \mathbb{R}^{k\times k \times d} $ is the visual feature for the $i$-th image $\bm{I}_i$.

With all the extracted features $\{\bm v_i\}_{i=1}^b$, we add them together as the inputs of the disease classification task, which is further used to learn the disease type representation as follows,
\vspace{-0.1in}
\begin{equation}\label{eq:disease_representation}
	\bm c_{pred} = \mathbf{W}_{cls} (\sum_{i=1}^b \text{AvgPool}(\bm v_i)) + \mathbf{b}_{cls},
	\vspace{-0.1in}
\end{equation}
where $\mathbf{W}_{cls}\in \mathbb{R}^{c\times d}$ and $ \mathbf{b}_{cls}\in \mathbb{R}^{c}$ are the weight and bias terms of a linear model, $\text{AvgPool}$ is the operation of average pooling, $c$ is the number of disease classes, and $\bm c_{pred}\in \mathbb{R}^c$ can be used to compute disease probabilities as a multi-label classification task with a sigmoid function, i.e., $p_{dc} = \text{sigmoid}(\bm c_{pred})$.

\subsubsection{Image-report matching} 
The next training task for VLR is to predict whether an image-report pair belongs to the same subject. In this subtask, after obtaining the image features $\{\bm{v}_i\}_{i=1}^b$ and the disease type representation $\bm c_{pred}$, we extract a context visual vector $\bm v$ by the pathological attention.

First, for each image feature $\bm v_i$, we use the disease type representation $\bm c_{pred}$ to learn the spatial attention score through a linear transformation,
\begin{align}
\label{eq:visual attn}
    \bm a_v =\mathbf{W}_{a} \text{tanh}(\mathbf{W}_{v} \bm v_i + \mathbf{W}_{c} \bm c_{pred})
\end{align}
where $\bm a_v\in\mathbb{R}^{k\times k}$, $ \mathbf{W}_{a}$, $ \mathbf{W}_{v}$ and $ \mathbf{W}_{c}$ are the linear transformation matrices. After that, we use the normalized spatial attention score $\bm \alpha_v = \text{softmax}(\bm a_v)$ to add visual features over all locations $(x,y)$ across the feature map,
\begin{align}
	\label{eq:visual attn agg}
	\bm v'_i =  \sum_{\forall x,y}  \bm \alpha_v(x,y)\bm v_i(x,y).
\end{align}

Then, we compute the context vector $\bm v$ of the input image set  $\{\bm{I}_i\}_{i=1}^b$ using a linear layer on the concatenation of all the representation $\bm v'_i$, $\bm v =\text{concat}(\bm v'_1,\cdots,\bm v'_b)\mathbf{W}_{f}$,
where $\mathbf{W}_{f} \in \mathbb{R}^{{bd \times d}}$ is the learnable parameter. 

For the image-report matching task, we also need a language representation, which is extracted by a BERT~\cite{DBLP:journals/corr/abs-1810-04805} model $f_l(\cdot)$ as the language encoder. $f_l(\cdot)$ converts the medical report $r$ into a semantic vector $\bm r = f_l(r) \in \mathbb{R}^{{d}}$. Finally, the probability of the input pair $(\{\bm I_i\}_{i=1}^b,r)$ coming from the same subject can be computed as
\begin{equation}
    p_{vl} = \text{sigmoid}(\bm r^T\bm v)\label{eq:vlm}.
\end{equation}

Given these two sub-tasks, we simultaneously optimize the cross-entropy losses for both disease classification and image-report matching to train the VLR module.

\subsection{LLR module pretraining}\label{llr model pretraining}
A medical report usually has some logical characteristics such as describing the patient’s medical images in a from-top-to-bottom order. Besides, the preceding and following sentences in a medical report may provide explanations for the same object or concept, or they may have certain juxtaposition, transition and progressive relations. Automatically learning such characteristics should be helpful for {\name} to generate high-quality medical reports. Towards this end, we propose to pretrain a language-language retrieval (LLR) module to search for the most relevant sentences for the next sentence generation. In particular, we introduce a self-supervised pretraining task for LLR to determine if two sentences $\{s_i,s_j\}$ come from the same report, i.e., \textbf{sentence-sentence matching}.

Similar to the VLR module, we use a BERT model $f_s(\cdot)$ as the sentence encoder to embed the sentence inputs $\{s_i,s_j\}$ into feature vectors $\bm s_i=f_s(s_i), \bm s_j = f_s(s_j)$.
Then the probability that two sentences $\{s_i,s_j\}$ come from the same medical report is measure by
\vspace{-0.05in}
\begin{equation}
    p_{ll} = \text{sigmoid}(\bm s_i^{\rm T} \bm s_j). \label{eq:llm}
    \vspace{-0.05in}
\end{equation}
Again, the cross-entropy loss is used to optimize the learning objective given probability $p_{ll}$ and the ground-truth label of whether $s_1$ and $s_2$ belong to the same medical report or not.


\subsection{Retrieval-based report generation}
Using the pretrained VLR and LLR modules, {\name} generates a medical report given a sequence of input images $\{\bm I_i\}_{i=1}^b$ using a novel hierarchical retrieval mechanism with a hierarchical language decoder. 


\subsubsection{VLR module for report-level retrieval}  

\paragraph{Report retrieval}
Let $\mathcal{D}^{(tr)}_r = \{r_j\}_{j=1}^{N_{tr}}$ denote the set of all the training reports, where $N_{tr}$ is the number of reports in the training dataset.
For each report $r_j$, {\name} first obtain its vector representation using $f_r(\cdot)$ in the VLR module, which is denoted as ${\bm r}_j=f_r(r_j)$. Let $\mathcal{P}_r = \{{\bm r}_j\}_{j=1}^{N_{tr}}$ denote the set of training report representations. Given the multi-modal medical images $\{\bm I_i\}_{i=1}^b$ of a subject, the VLR module aims to return the \textit{top $k_r$ medical reports} $\{r_j^\prime\}_{j=1}^{k_r}$ as well as \textit{medical keywords} within in the retrieved reports.


Specifically, {\name} extracts the image feature $\bm v$ for $\{\bm I_i\}_{i=1}^b$ using the pathological attention mechanism as described in Section \ref{vlrm pretraining}. According to Eq.~(\ref{eq:vlm}), {\name} then computes a image-report matching sore $p_{vl}$ between $\bm v$ and each $\bm r \in \mathcal{P}_r$.
The top $k_r$ reports $\{r_j^\prime\}_{j=1}^{k_r}$ with the largest scores $p_{vl}$ are considered as the most relevant medical reports corresponding to the images, and they are selected as the template descriptions. 
From these templates, we identify $n$ medical keywords $\{w_i\}_{i=1}^n$ using a dictionary as a summarization of the template information. The medical keyword dictionary includes disease phenotype, human organ, and tissue, which consists of 36 medical keywords extracted from the training data with the highest frequency. 


\paragraph{Report template representation learning} 
The retrieved reports are highly related to the given images, which should be helpful for the report generation. To make full use of them, we need to learn a report template representation using the image feature $\bm v$, the features of retrieved reports $\{{\bm r}^\prime_j\}_{j=1}^{k_r}$, medical keywords embeddings $\{\bm w_i\}_{i = 1}^n$ for $\{w_i\}_{i=1}^n$ learned from the pretrained word embeddings, and the disease embeddings $\{\bm c_k\}_{k = 1}^m$ from predicted disease labels $\{ c_k\}_{k = 1}^m$ using Disease Classification in Section~\ref{disease classifition}.

We propose a new \textbf{multi-query attention} mechanism to learn the report template representation. To specify, we use the image features $\bm v$ as the key vector $\bm K$, the retrieved report features $\{{\bm r}^\prime_j\}_{j=1}^{k_r}$ as the value matrix $\bm V$, and the embeddings of both medical keywords $\{\bm w_i\}_{i = 1}^n$ and disease labels $\{\bm c_k\}_{k = 1}^m$ as the query vectors $\bm Q$. We modify the original self-attention~\cite{vaswani2017attention} into a multi-query attention. For each query vector $\bm Q_i$ in $\bm Q$, we first get a corresponding attended feature and then transform them into the \textbf{report template vector} ${\bm r}_s$ after concatenation,
\begin{equation}
	\begin{split}
		{\bm r}_s &= \text{MultiQuery}(\{{\bm Q_i}\}_{i=1}^{n}, {\bm K}, {\bm V}) \\ 
		&= \text{concat}(\text{attn}_1, \cdots, \text{attn}_n) {\bm W^O},
		\end{split}
	\label{eq:query}
\end{equation}
where $\text{attn}_i = \text{Attention}({\bm Q}_i, {\bm K}{\bm W^K}, {\bm V}{\bm W^V })$, and ${\bm W^K}$, ${\bm W^V}$ and ${\bm W^O}$ are the transformation matrices. Generally, the Attention function is calculated by
\vspace{-0.1in}
\begin{equation*}\small
	 \text{Attention}({\bm Q_g}, {\bm K_g}, {\bm V_g}) = \text{softmax}(\frac{{\bm Q_g}{\bm K_g}^{\rm T}}{\sqrt{d_g}}) {\bm V_g},
	 \vspace{-0.05in}
\end{equation*}
where ${\bm Q}, {\bm K}, {\bm V}$ are queries, keys and values in general case, and $d_g$ is the dimension of the query vector.

\subsubsection{LLR module for sentence-level retrieval} 

Since retrieved reports $\{r_j^t\}_{j=1}^{k_r}$ are highly associated with the input images, the sentence within those reports must contain some instructive pathological information that is helpful for sentence-level generation. Towards this end, we first select sentences from the retrieved reports and then learn sentence-level template representation.

\paragraph{Sentence retrieval} 
We first divide the retrieved reports into $L$ candidate sentences $\{s_j\}_{j=1}^{L}$ as the retrieval pool in the LLR module. Given the pretrained LLR language encoder $f_s(\cdot)$, we can obtain the sentence-level feature pool, which is $\mathcal{P}_s = \{f_s(s_j)\}_{j=1}^{L} =\{\bm s_j\}_{j=1}^{L}$. 
Assume that the generated sentence at time $t$ is denoted as $o_{t}$, and its embedding is $\bm o_{t} = f_s(o_{t})$, which is used to find $k_s$ sentences $\{s_{j}^{\prime}\}_{j=1}^{k_s}$ with the highest probabilities $p_{ll}$ from the candidate sentence pool using Eq.~(\ref{eq:llm}) in Section~\ref{llr model pretraining}. 


\paragraph{Sentence template representation learning} 
Similar to the report template representation, we still use the multi-query attention mechanism.
From the retrieved $k_s$ sentences, we extract the medical keywords $\{w^\prime_i\}_{i=1}^{n}$. 
Besides, we have the predicted disease labels $\{c_k\}_{k = 1}^m$. Their embeddings are considered as the query vectors.
The embeddings of the extracted sentence, i.e., $\{f_s(s_{j}^{\prime})\}_{j=1}^{k_s} = \{\bm s_{j}^{\prime}\}_{j=1}^{k_s}$, are treated as the value vectors. 
The key vector is the current sentence~(word) hidden state $\bm h_t^s$~($\bm h_i^w$), which will be introduced in Section~\ref{Hierarchical Language Decoder}. 
According to Eq.~(\ref{eq:query}), we can obtain the sentence template representation at time $t$, which is denoted as $\bm u_t$ ($\bm u_i^w$ used for word-level generation).


\subsubsection{Hierarchical language decoder} \label{Hierarchical Language Decoder}
With the extracted features by the retrieval mechanism described above, we apply a hierarchical decoder to generate radiology reports according to the hierarchical linguistics structure of the medical reports. The decoder contains two layers, i.e., a sentence LSTM decoder that outputs sentence hidden states, and a word LSTM decoder which decodes the sentence hidden states into natural languages. In this way, reports are generated sentence by sentence. 
\paragraph{Sentence-level LSTM}
For generating the $t$-th sentence, {\name} first uses the previous $t-1$ sentences to learn the sentence-level hidden state $\bm h_t^s$. Specifically, {\name} learns the image feature $\bm v^{s}$ based on Eq.~(\ref{eq:visual attn agg}). When calculating the attention score with Eq.~(\ref{eq:visual attn}), we consider both the information obtained from the previous $t-1$ sentences (the hidden state $\bm h_{t-1}^s$) and the predicted disease representation from Eq.~(\ref{eq:disease_representation}), i.e., replacing $\bm c_{pred}$ with $\text{concat}(\bm h_{t-1},\bm c_{pred})$. Then the concatenation of the image feature $\bm v^{s}$, the report template representation $\bm r_s$ from Eq.~(\ref{eq:query}), and the sentence template representation $\bm u_{t-1}^s$ is used as the input of the sentence LSTM to learn the hidden state $\bm h_t^s$
\begin{equation}\label{eq:lstm s}
    \bm h_t^s = \text{LSTM}_s(\text{concat}(\bm v^{s}, \bm u_{t-1}^s, \bm r_s), \bm h_{t-1}^s),
\end{equation}
where $\bm u_{t-1}^s$ is obtained using the multi-query attention, the key vector is the hidden state $\bm h_{t-1}^s$, the value vectors are the representations of the retrieved sentences according to the $(t-1)$-th sentence, and the query vectors are the embeddings of both medical keywords extracted from the retrieved sentences and the predicted disease labels.

\paragraph{Word-level LSTM}
Based on the learned $\bm h_t^s$, {\name} conducts the word-by-word generation using a word-level LSTM. For generating the $(i+1)$-th word, {\name} first learns the image feature $\bm v^{w}$ using Eq.~(\ref{eq:visual attn}) by replacing $\bm c_{pred}$ with $\bm h_i^w$ in Eq.~(\ref{eq:visual attn}), where  $\bm h_i^w$ is the hidden state of the $i$-th word. {\name} then learns the sentence template representation $\bm u_{i}^w$ using the multi-query attention, where the key vector is the hidden state $\bm h_i^w$, value and query vectors are the same as those used for calculating $\bm u_{t-1}^s$. Finally, the concatenation of $\bm h_t^s$, $\bm u_{i}^w$, $\bm v^{w}$, and $\bm r_s$ is taken as the input of the word-level LSTM to generate the $(i+1)$-th word as follows:
\begin{equation}\small
\begin{split} \label{eq:lstm w}
    &\bm h_i^w = \text{LSTM}_w(\text{concat}(\bm h_t^s, \bm u_{i}^w, \bm v^{w}, \bm r_s), \bm h_{i-1}^w),\\
    &w_{i+1} = \textrm{argmax}(\text{softmax}(FFN(\bm h_i^w ))),
\end{split}
\end{equation}
where $FFN(\cdot)$ is the feed-forward network.

Note that for the first sentence generation, we set $\bm u_{0}$ as $\textbf{0}$, and $\bm h_{0}$ is the randomly initialized vector, to learn the sentence-level hidden state $\bm h^s_1$. When generating the words of the first sentence, we set $\bm u_{i}^w$ as the $\textbf{0}$ vector.

\section{Experiments}
\subsection{Datasets and baselines}
\paragraph{Datasets}
\textbf{Open-i}\footnote{\url{https://openi.nlm.nih.gov/faq\#collection}}~\cite{demner2016preparing} (\textit{a.k.a} IU X-Ray) provides 7,470 chest X-rays with 3,955 radiology reports. In our experiments, we only utilize samples with both frontal and lateral views, and with complete findings and impression sections in the reports. This results in totally 2,902 cases and 5,804 images. 
\textbf{MIMIC-CXR}\footnote{\url{https://physionet.org/content/mimic-cxr/2.0.0/}}~\cite{johnson2019mimic} contains 377,110 chest X-rays associated with 227,827 radiology reports, divided into subsets. We use the same criterion to select samples, which results in 71,386 reports and 142,772 images. 

For both datasets, we tokenize all words with more than 3 occurrences and obtain 1,252 tokens on the Open-i dataset and 4,073 tokens on the MIMIC-CXR dataset, including four special tokens $\langle\textsc{pad}\rangle$, $\langle\textsc{start}\rangle$, $\langle\textsc{end}\rangle$, and $\langle\textsc{unk}\rangle$. The findings and impression sections are concatenated as the ground-truth reports. We randomly divide the whole datasets into train/validation/test sets with a ratio of 0.7/0.1/0.2. To conduct the disease classification task, we include 20 most frequent finding keywords extracted from MeSH tags as disease categories on the Open-i dataset and 14 CheXpert categories on the MIMIC-CXR dataset.

\paragraph{Baselines}
On both datasets, we compare with four state-of-the-art image captioning models: CNN-RNN~\cite{vinyals2015show}, CoAttn~\cite{jing2017automatic}, MvH+AttL~\cite{yuan2019automatic}, and V-L Retrieval. V-L Retrieval only uses the retrieved report templates with the highest probability as prediction without the generation part based on our pretrained VLR module.
Due to the lack of the opensource code for~\cite{wang2018tienet,li2019knowledge,li2018hybrid,donahue2015long} and the template databases for~\cite{li2019knowledge,li2018hybrid}, we only include the \textit{reported results} on the Open-i dataset in our experiments.

\begin{table*}[]
    \begin{center}
    \scriptsize
    \begin{tabular}{c|c|l|c c c c c c ||c}
    \hline
        Dataset  & Type & Model & CIDEr & ROUGE-L & BLEU-1 & BLEU-2 & BLEU-3 & BLEU-4 & AUC\\
        \hline
        \multirow{10}{*}{\textbf{Open-i}} & \multirow{5}{*}{Generation}
        & CNN-RNN~\cite{vinyals2015show} & 0.294 & 0.307 & 0.216 & 0.124 & 0.087 & 0.066 & 0.426\\
        & & LRCN~\cite{donahue2015long}* & 0.285 & 0.307 & 0.223 & 0.128 & 0.089& 0.068 & --\\
        & & Tie-Net~\cite{wang2018tienet}* & 0.279 &  0.226 & 0.286 &  0.160 & 0.104&  0.074 & --\\
        & & CoAtt~\cite{jing2017automatic} & 0.277 & 0.369 & 0.455 & 0.288 & 0.205& 0.154& 0.707\\
        & & MvH+AttL~\cite{yuan2019automatic} &0.229  & 0.351 & 0.452 & 0.311 &0.223 & 0.162 & 0.725\\\cline{2-10}
        & \multirow{4}{*}{Retrieval}
        & V-L Retrieval & 0.144 & 0.319 & 0.390 & 0.237 &0.154 & 0.105 & 0.634 \\
        & & HRGR-Agent~\cite{li2018hybrid}* & 0.343 & 0.322 & 0.438 & 0.298 & 0.208 & 0.151& --\\
        & & KERP~\cite{li2019knowledge}*  & 0.280 & 0.339 & \textbf{0.482} & 0.325 & 0.226& 0.162&-- \\
        & & {\name} & \textbf{0.345} & \textbf{0.382} & 0.471 & \textbf{0.336} & \textbf{0.238} & \textbf{0.166}& \textbf{0.814}\\
        \cline{2-10}
         &  \multicolumn{2}{c|}{Ground Truth} & -- & -- & -- & -- & -- & -- &0.915\\
         \hline \hline
         \multirow{6}{*}{\textbf{MIMIC-CXR}} & \multirow{3}{*}{Generation}
         & CNN-RNN~\cite{vinyals2015show} & 0.245 & 0.314 & 0.247 & 0.165 & 0.124 & 0.098& 0.472\\
         & & CoAtt~\cite{jing2017automatic} & 0.234 & 0.274 & 0.410 & 0.267 & 0.189& 0.144& 0.745\\
         & & MvH+AttL~\cite{yuan2019automatic} &0.264  & 0.309 &0.424 & 0.282 &0.203 & 0.153 & 0.738\\ \cline{2-10}
        & \multirow{2}{*}{Retrieval}
        & V-L Retrieval & 0.186 & 0.232 & 0.306 & 0.179 &0.116 & 0.076& 0.579\\
        &  &
         {\name} & \textbf{0.306} & 0.\textbf{332} & \textbf{0.438} & \textbf{0.297} & \textbf{0.216} & \textbf{0.164}& \textbf{0.833}\\\cline{2-10}
         &  \multicolumn{2}{c|}{Ground Truth} & -- & -- & -- & -- & -- & -- &0.923\\
         \hline
    \end{tabular}
    \end{center}
    \vspace{-0.15in}
    \caption{Automatic evaluation on the Open-i and MIMIC-CXR datasets. * indicates the results reported in~\cite{li2019knowledge}.}
    \label{tab:Automatic evaluation}
    \vspace{-0.2in}
\end{table*}

\subsection{Experimental setup}
All input images are resized to $512\times 512$, and the feature map from
DenseNet-121~\cite{huang2017densely} is $1024\times 16\times 16$. During training, we use random cropping and color histogram equalization for data augmentation.

To pretrain the VLR module, the maximum length of the report is restricted to 128 words. We train VLR module for 100 epochs with an Adam~\cite{kingma2014adam} optimizer with 1e-5 as the initial learning rate, 1e-5 for L2 regularization, and 16 as the mini-batch size. 
To pretrain the LLR module, the maximum length of each sentence is set to 32 words. We optimize the LLR module for 100 epochs with an Adam~\cite{kingma2014adam} optimizer with the initial learning rate of 1e-5 and a mini-batch size of 64. The learning rate is multiplied by 0.2 every 20 epochs. 

To train the full model for {\name}, we set the retrieved reports number $k_r=5$ and sentences number $k_s=5$. Extracting $n=5$ medical keywords and predicting $m=5$ disease labels are used for report generation. Both sentence and word LSTM have 512 hidden units. We freeze the weights for the pretrained VLR and LLR modules and only optimize on the language decoder. We set the initial learning rate as 3e-4 and mini-batch size as 32. {\name} takes 10 hours to train on the Open-i dataset and 3 days on the MIMIC-CXR dataset with four GeForce GTX 1080 Ti GPUs. 

\subsection{Quantitative and qualitative results}
\tablename~\ref{tab:Automatic evaluation} shows the CIDEr, ROUGE-L, BLUE, and AUC scores achieved by different methods on the test sets of Open-i and MIMIC-CXR. 

\paragraph{Language evaluation}
From \tablename~\ref{tab:Automatic evaluation}, we make the following observations. First, compared with \textit{Generation}-based model, \textit{Retrieval}-based model that uses the template reports as results has set up a relatively strong baseline for medical report generation. 
Second, compared with V-L retrieval, other \textit{Retrieval}-based approaches perform much better in terms of all the metrics. This again shows that that by integrating the information retrieval method into the deep sequence generation framework, we can not only use the retrieved language information as templates to help generate long sentences, but also overcome the monotony of only using the templates as the generations.
Finally, we see that the proposed {\name} achieves the highest language scores on 5/6 metrics on Open-i datasets and all metrics on MIMIC-CXR among all methods. {\name} not only improves current SOTA model CoAttn~\cite{jing2017automatic} by 5\% and MvH+AttL~\cite{yuan2019automatic} by 4\% on Open-i in average, but also goes beyond SOAT retrieval-based approaches like KERP~\cite{li2019knowledge} and HRGR-Agent~\cite{li2018hybrid} and significantly improves the performance, even \textit{without using manually curated template databases}. This illustrates the effectiveness of automatically learning templates and adopting hierarchical retrieval in writing medical reports.

\paragraph{Clinical evaluation} We train two report classification BERT models on both datasets and use it to judge whether the generated reports correctly reflect the ground-truth findings. We show the mean ROC-AUC scores achieved by generated reports from different baselines in the last column of Table~\ref{tab:Automatic evaluation}. We can observe that {\name} achieves the highest AUC scores compared with other baselines. In addition, our method achieves the AUC scores that are very close to those of professional doctors' reports, with 0.814/0.915 and 0.833/0.923 on two datasets. This shows that the generation performance of {\name} has approached the level of human domain experts, and it embraces great medical potentials in identifying disease-related medical findings.

\paragraph{Human evaluation}
We also qualitatively evaluate the quality of the generated reports via a user study. 
We randomly select 50 samples from the Open-i test set and collect ground-truth reports and the generated reports from both MvH+AttL~\cite{yuan2019automatic} and {\name} to conduct the human evaluation.
Two experienced radiologists were asked to give ratings for each selected report, in terms of whether the generated reports are realistic and relevant to the X-ray images. 
The ratings are integers from one to five. The higher, the better. 

\tablename~\ref{tab:user study} shows average human evaluation results on {\name}
compared with Ground Truth reports and generations of MvH+AttL~\cite{yuan2019automatic} on Open-i, evaluated in terms of realistic scores and relevant scores. {\name} achieves
much higher human preference than the baseline model, even approaching the performance of Ground Truth reports that wrote by experienced radiologists. It shows that {\name} is able to generate accurate clinical reports that are comparable to domain experts.
\begin{table}[!htb]
    \begin{center}
    \scriptsize
    \begin{tabular}{c|cc}
    \hline
         Method &  Realistic Score & Relevant Score  \\	
        \hline
        Ground Truth & 3.85 & 3.82\\ 					
        MvH+AttL~\cite{yuan2019automatic} & 2.50& 2.57\\
        {\name} &3.68 & 3.44\\\hline
    \end{tabular}
    \end{center}
    \vspace{-0.15in}
    \caption{User study conducted by two domain experts. }
    \label{tab:user study}
    \vspace{-5mm}
\end{table}

\begin{figure*}
    \begin{center}
    \scriptsize
    \begin{tabular}{m{0.11\textwidth} m{0.12\textwidth} p{0.22\textwidth}p{0.18\textwidth}p{0.2\textwidth}}

       \textbf{Frontal Image} & \textbf{Lateral Image}  & \textbf{Ground Truth} & \textbf{MvH+AttL} & \textbf{{\name}}   \\\hline
 \includegraphics[width=0.82in,height=0.82in]{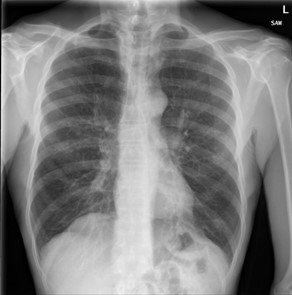} &  \includegraphics[width=0.82in,height=0.82in]{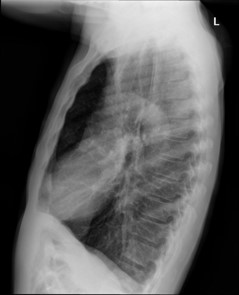} 
 & \pbox{3.5cm}{\textcolor{red}{emphysematous changes}. resolution of prior \textcolor{red}{right midlung} infiltrate. previous $\langle\textsc{unk}\rangle$ is normal in size and contour. lungs are clear. \textbf{no focal consolidation pneumothorax or pleural effusion.} interval $\langle\textsc{unk}\rangle$ of previously described right midlung opacity suggesting resolved $\langle\textsc{unk}\rangle$ process. \textcolor{red}{lungs are hyperexpanded} with flattened diaphragms. $\langle\textsc{unk}\rangle$ and soft tissue are unremarkable. 
}   
         &  \pbox{3cm}{no acute cardiopulmonary disease. the heart is normal in size. the lungs are clear. \textbf{there is no pleural effusion or pneumothorax.} of the right clavicle. $\langle\textsc{unk}\rangle$ $\langle\textsc{unk}\rangle$ are present. $\langle\textsc{unk}\rangle$ to the glenoid joints.  
} 
        & \pbox{3cm}{\textcolor{red}{hyperexpanded lungs}. $\langle\textsc{unk}\rangle$ \textcolor{red}{right upper lobe} $\langle\textsc{unk}\rangle$. no focal pneumonia. the cardiomediastinal silhouette is normal in size and contour. \textbf{negative for focal consolidation pneumothorax or large pleural effusion.}  negative for acute bone abnormality.  
}
 \\\hline
 
        \includegraphics[width=0.82in,height=0.82in]{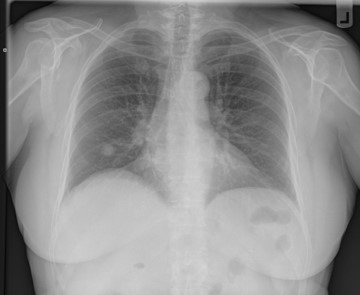} &  \includegraphics[width=0.82in,height=0.82in]{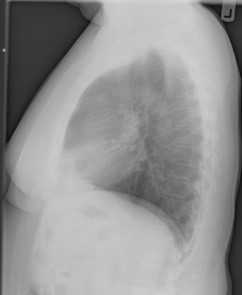} &  \pbox{3.5cm}{chest. \textcolor{red}{large nodule at the right lung base} that probably represents a granuloma although not it is not $\langle\textsc{unk}\rangle$ calcified. \textcolor{red}{there is a $\langle\textsc{unk}\rangle$ mm nodule in the right lower lobe} that is relatively dense but not $\langle\textsc{unk}\rangle$ calcified on the corresponding rib series. there are probably right hilar calcified lymph $\langle\textsc{unk}\rangle$. lungs otherwise are clear. \textbf{there is no pleural effusion}. left ribs. \textit{no fracture or focal bony destruction.} } 
        & \pbox{3cm}{no acute cardiopulmonary disease. the heart is normal in size and contour. are clear without evidence of infiltrate. \textbf{is no pneumothorax}. \textit{degenerative changes of the thoracic spine.}. head.. } 
        &\pbox{3cm}{right upper lobe pneumonia. consideration may be given for primary or $\langle\textsc{unk}\rangle$. \textcolor{blue}{recommend ct of the chest may be helpful for further diagnosis.} in the interval a \textcolor{red}{3 cm $\langle\textsc{unk}\rangle$ mass has developed in the right lower lobe}. \textbf{no pneumothorax or pleural effusion}. the mediastinal contours are normal. 
}  
  \\\hline

    \end{tabular}
    \end{center}
    \vspace{-0.15in}
    \caption{Examples of ground-truth and generated reports by MvH+AttL~\cite{yuan2019automatic} and {\name}. Highlighted \textcolor{red}{red} phrases are medical abnormality terms that generated and ground-truth reports have in common. \textbf{Bold} terms are common descriptions of normal tissues. The text in \textit{italics} is the opposite meaning of the generated report and the actual report. We mark the supplementary comments to the original report in \textcolor{blue}{blue}.}
    \label{fig:qualitative}
\end{figure*}

\begin{table*}[!htb]
    \begin{center}
    \footnotesize
    \begin{tabular}{c|l|c c c c c c}
    \hline
        Dataset &  Model & CIDEr & ROUGE-L & BLEU-1 & BLEU-2 & BLEU-3 & BLEU-4 \\
        \hline
        \multirow{4}{*}{\textbf{Open-i}}  &{\name} w/o VLRM & 0.333 & 0.373 & 0.466 & 0.324 & 0.229 & 0.159\\
        &{\name} w/o LLRM &  0.329 & 0.354 & 0.453 & 0.307 & 0.215 & 0.154\\
        &{\name} w/o HLD &0.284 & 0.317 & 0.434 & 0.295 & 0.208 & 0.149\\
         & {\name} & \textbf{0.345} & \textbf{0.382} & \textbf{0.471} & \textbf{0.336} & \textbf{0.238} & \textbf{0.166}\\
         \hline \hline
         \multirow{5}{*}{\textbf{MIMIC-CXR}} &{\name} w/o VLRM & 0.294 & 0.317 & 0.432 & 0.288 & 0.209 & 0.161\\
        &{\name} w/o LLRM & 0.283 & 0.305 & 0.425 & 0.280 & 0.204 & 0.157\\
        &{\name} w/o HLD & 0.263 & 0.287 & 0.418 & 0.265 & 0.187 & 0.146\\
        & {\name} & \textbf{0.306} & 0.\textbf{332} & \textbf{0.438} & \textbf{0.297} & \textbf{0.216} & \textbf{0.164}\\
         \hline
    \end{tabular}
    \end{center}
    \vspace{-0.15in}
    \caption{Ablation study on both Open-i and MIMIC-CXR datasets.}
    \label{tab:Ablation study}
    \vspace{-6mm}
\end{table*}

\paragraph{Qualitative analysis}
Figure~\ref{fig:qualitative} shows qualitative results of {\name} and baseline models on the Open-i dataset. {\name} not only produces longer reports compared with MvH+AttL but also accurately detects the medical findings in the images~(marked in \textcolor{red}{red} and \textbf{bold}). On the other hand, we find that {\name} is able to put forward some supplementary suggestions~(marked in \textcolor{blue}{blue}) and descriptions, which are not in the original report but have diagnostic value. The underlying reason for this merit comes from the memory retrieval mechanism that introduces prior medical knowledge to facilitate the generation process.


\subsection{Ablation study}
We perform ablation studies on the Open-i and MIMIC-CXR datasets to investigate the effectiveness of each module in {\name}. In each of the following studies, we change one module with other modules intact.

\vspace{-0.05in}
\paragraph{Removing the VLR module}
In this experiment, global report feature $\bm r_s$ is neglected in Eqs.~(\ref{eq:lstm s}) and~(\ref{eq:lstm w}), and the first sentence is generated only based on image features. The LLR module keeps its functionality. However, instead of looking for sentence-level templates from the retrieved reports, it searches for most relevant sentences from \textit{all the reports}. As can be seen from \tablename~\ref{tab:Ablation study}, removing VLR module (``w/o VLRM'') leads to performance reduction by 2\% on average. This demonstrates that visual-language retrieval is capable in sketching out the linguistic structure of the whole report. The rest of the language generation is largely influenced by report-level context information.

\vspace{-0.05in}
\paragraph{Removing the LLR module}
The generation of $(t+1)$-th sentence is based on the global report feature $\bm r_s$ and the image feature $\bm v$, without using the retrieved sentences information in Eq.~(\ref{eq:lstm w}). \tablename~\ref{tab:Ablation study} shows that removing LLR module (``w/o LLRM'') results in the decease of average evaluation scores by 4\% compared with the full model. This verifies that the LLR module plays an essential role in generating long and coherent clinical reports.

\vspace{-0.05in}
\paragraph{Replacing hierarchical language decoder}
We use a single layer LSTM that treats the whole report as a long sentence and conduct the generation word-by-word. \tablename~\ref{tab:Ablation study} shows that replacing hierarchical language decoder with a single-layer LSTM~(``w/o HLD'') introduces dramatic performance reduction. This phenomenon shows that the hierarchical generative model can effectively and greatly improve the performance of long text generation tasks.

\section{Conclusions}
\vspace{-0.1in}
Automatically generating accurate reports from medical images is a key challenge in medical image analysis. In this paper, we propose a novel model named {\name} to solve this problem based on hierarchical retrieval techniques. In particular, {\name} consists of three main modules, which are the visual-language retrieval (VLR) module, the language-language retrieval (LLR) module, and the hierarchical language decoder. These three modules tightly work with each other to automatically generate medical reports. Experimental results on two datasets demonstrate the effectiveness of the proposed {\name}. Besides, qualitative studies show that {\name} is able to generate meaningful and realistic medical reports.

\bibliographystyle{acl_natbib}
\bibliography{acl2021}


\end{document}


\maketitle
This supplement gives details and results omitted in the main paper.
\begin{itemize}
    \item Section 1: key notations used in this paper.
    \item Section 2: Detailed experiments setups, and additional experiments for model analysis.
\end{itemize}

\section{Notations}
Table~\ref{tab:notations} lists the key notations used in this paper.
\begin{table}[H]
    \centering
    \scriptsize
    \begin{tabular}{l |l}
    \hline
    \textbf{Notation} & \textbf{Description}\\\hline
        $\bm I$ &  Image\\
         $\bm v$ & Visual feature \\
        $r$/$r_j$ & Medical report/the $j$-th report\\
        $\bm r$/$\bm r_j$ & Representation of a medical report/the $j$-th report\\
        $\bm r_s$ & Report-level template feature \\
        $r_j^\prime$ & The $j$-th retrieved report\\
        $\bm r_j^\prime$ & Representation of the $j$-th retrieved report\\
         $s_j$ & The $j$-th sentence\\
         $\bm s_j$ & Representation of the $j$-th sentence\\
         $s_j^\prime$ & The $j$-th sentence in the retrieved report set\\
         $\bm s_j^\prime$ & Representation of the $j$-th retrieved sentence\\
          $\bm u_t$ & Sentence-level template feature \\
         $o_t$ & Generated sentence at step $t$\\
          $\bm o_t$ & Sentence-level feature generated sentence at step $t$\\
         $w$ & Medical keywords\\
          $\bm w$ & Medical keywords embedding\\
          $c$ & Predicted disease label\\
          $\bm c$ & Predicted disease embedding\\
        $f_v(\cdot)$ & Image Endocer \\
        $f_l(\cdot)$ & Report Language Endocer \\
        $f_s(\cdot)$ & Sentence Language Endocer \\
        $\mathbf{W}$ & Learnable weights for linear model\\
        $\mathbf{b}$ & Learnable bias for linear model\\
        $\bm h^s_t$ & Hidden state for sentence LSTM at step $t$\\
         $\bm h^w_i$ & Hidden state for word LSTM at step $t$\\
         $\bm a$/$\bm \alpha$ & Attention score/Normalized attention score\\
         $k_r$ & Number of retrieved reports\\
         $k_s$ & Number of retrieved sentences\\
         \hline
    \end{tabular}
    \caption{Key notations}
    \label{tab:notations}
\end{table}


\section{Experiments}
In this section, we first refine the experimental setup and design details, including medical keywords and hyperparameter selections. We visualize more experimental results on the Open-i dataset, including (1) image spatial attention, (2) multi-query attention, (3) retrieved report/sentence templates, and (4) generated reports by {\name}. 

\subsection{Medical keywords}
In Section 3, when calculating both report-level and sentence-level template representations, we extract medical keywords from the retrieved reports or sentences. The medical keyword dictionary includes diseasephenotype, human organ, and tissue, which consists of 36 medical keywords extracted from the training data with the highest frequency as shown in Table~\ref{tab:Medical keywords}.

\begin{table}[!htb]
    \tiny
    \centering
    \begin{tabular}{|c|c|c|}
    \hline
      normal & cardiomegaly& scoliosis\\ degenerative& fractures& bone\\
    pleural & effusion & thickening \\ pneumothorax & hernia hiatal &calcinosis \\
    emphysema&pulmonary&pneumonia\\infiltrate&consolidation&pulmonary\\
    edema&atelectasis&cicatrix\\opacity&nodule&mass\\
    airspace&hypoinflation&hyperdistention\\pneumonia&granuloma&scoliosis\\
    inflation&surgical&diffuse\\fibrosis&cardiac&blunting   \\\hline
    \end{tabular}
    \caption{Selected medical keywords.}
    \label{tab:Medical keywords}
\end{table}

\subsection{Experimental Setting}
The hyperparameter settings for training the VLR module, LLR module and full {\name} are shown in Table~\ref{tab:VLRM exp setting}, Table~\ref{tab:LLRM exp setting}, and Table~\ref{tab:full exp setting}, respectively. 

\begin{table}
    \centering
    \caption{Hyperparameter settings for training Visual-Language Retrieval~(VLR) Module}
    \label{tab:VLRM exp setting}
    \tiny
    \begin{tabular}{l|c|l}
    \hline
         \textbf{Name} & \textbf{Value} & \textbf{Description} \\
         \hline
         Manual seed & 123 & The manual random seed for pytorch \\\hline
         Input size & $512\times512$ & Training image size\\\hline
         Maximum report length & $128$ & Maximum words in a report\\\hline
         Optimizer & Adam & \parbox{15em}{Optimizers for the image encoder and language encoder}\\\hline
         Image learning rate & \num{1e-4} & Multiply by 0.1 at the 50-th epoch \\\hline
         Language learning rate & \num{1e-5} & Multiply by 0.1 at the 50-th epoch \\\hline
         L2 regularization & \num{1e-5} & Ratio for  L2 regularization\\\hline
         Gradient clip & 5.0 & Clips gradient magnitude under 5.0\\\hline
         Betas $(\beta_1,\beta_2)$ &(0.9, 0.999)& \parbox{15em}{Adam coefficients used for computing running averages of gradient and its square}\\\hline
        Train epoch & 100 & -\\\hline
        Batch size & 16 &-\\
        \hline
    \end{tabular}
\end{table}
\begin{table}
    \centering
    \tiny
    \caption{Hyperparameter settings for training Language-Language Retrieval Module}
    \label{tab:LLRM exp setting}
    \begin{tabular}{l|c|l}
    \hline
        \textbf{Name} & \textbf{Value} & \textbf{Description} \\
         \hline
         Manual seed & 123 & The manual random seed for pytorch \\\hline
          Maximum sentence length & $32$ & Maximum words in a sentence\\\hline
         Optimizer & Adam & \parbox{15em}{Optimizers for the language encoder}\\\hline
         Learning rate & \num{1e-5} & Multiply by 0.2 at every 20 epochs \\\hline
         Betas $(\beta_1,\beta_2)$ &(0.9, 0.999)& \parbox{15em}{Adam coefficients used for computing running averages of gradient and its square}\\\hline
        Train epoch & 100 & -\\\hline
        Batch size & 64 &-\\
        \hline
    \end{tabular}
\end{table}
\begin{table}
    \centering
    \tiny
    \caption{Hyperparameter settings for training {\name}}
    \label{tab:full exp setting}
    \begin{tabular}{l|c|l}
    \hline
         \textbf{Name} & \textbf{Value} & \textbf{Description} \\
         \hline
         Manual seed & 123 & The manual random seed for pytorch \\\hline
         Input size & $512\times512$ & Training image size\\\hline
         Maximum sentence length & $32$ &  \parbox{15em}{Maximum word number in a sentence for sentence retrieval pool}\\\hline
         Maximum report  length& $128$ & \parbox{15em}{Maximum word number in a report for report retrieval pool}\\\hline
         Optimizer & Adam & \parbox{15em}{Optimizers for the language decoder~(other weights fixed)}\\\hline
         Learning rate & \num{3e-4} & \parbox{15em}{Multiply by 0.1 at the [50]-th epoch} \\\hline
         L2 regularization & \num{1e-5} & Ratio for  L2 regularization\\\hline
         Gradient clip & 5.0 & Clips gradient magnitude under 5.0\\\hline
         Betas $(\beta_1,\beta_2)$ &(0.9, 0.999)& \parbox{15em}{Adam coefficients used for computing running averages of gradient and its square}\\\hline
        Train epoch & 100 & -\\\hline
        Batch size & 32 &-\\\hline
        Retrieved report number &5&-\\\hline
        Retrieved sentence number &5&-\\
        \hline
    \end{tabular}
\end{table}

\subsection{Image pathological spatial attention}

\begin{figure*}[htb]
    \centering
    \tiny
        \begin{tabular}{m{0.3\textwidth} m{0.3\textwidth}m{0.3\textwidth} }
            \includegraphics[width=2in]{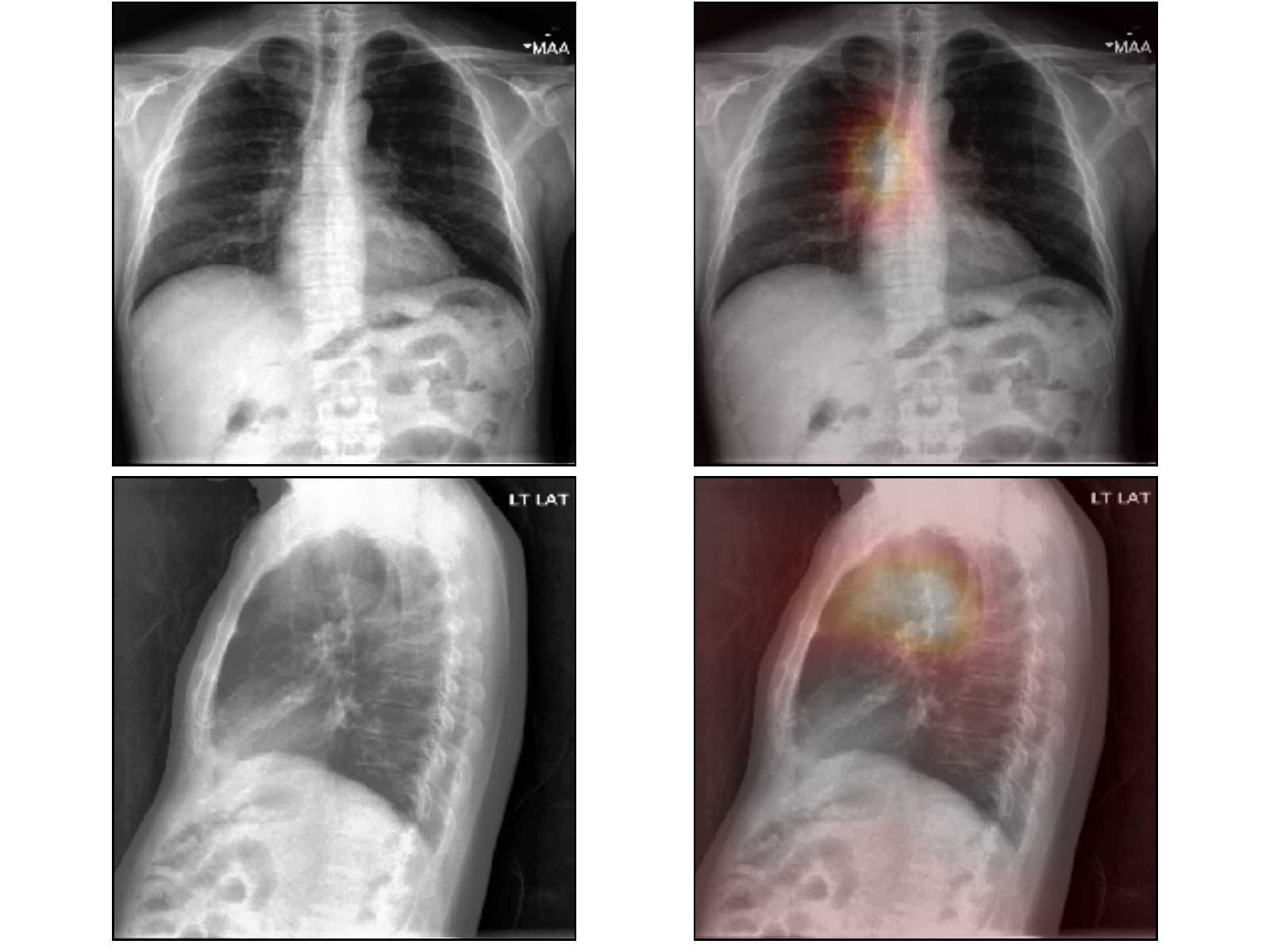} &
     \includegraphics[width=2in]{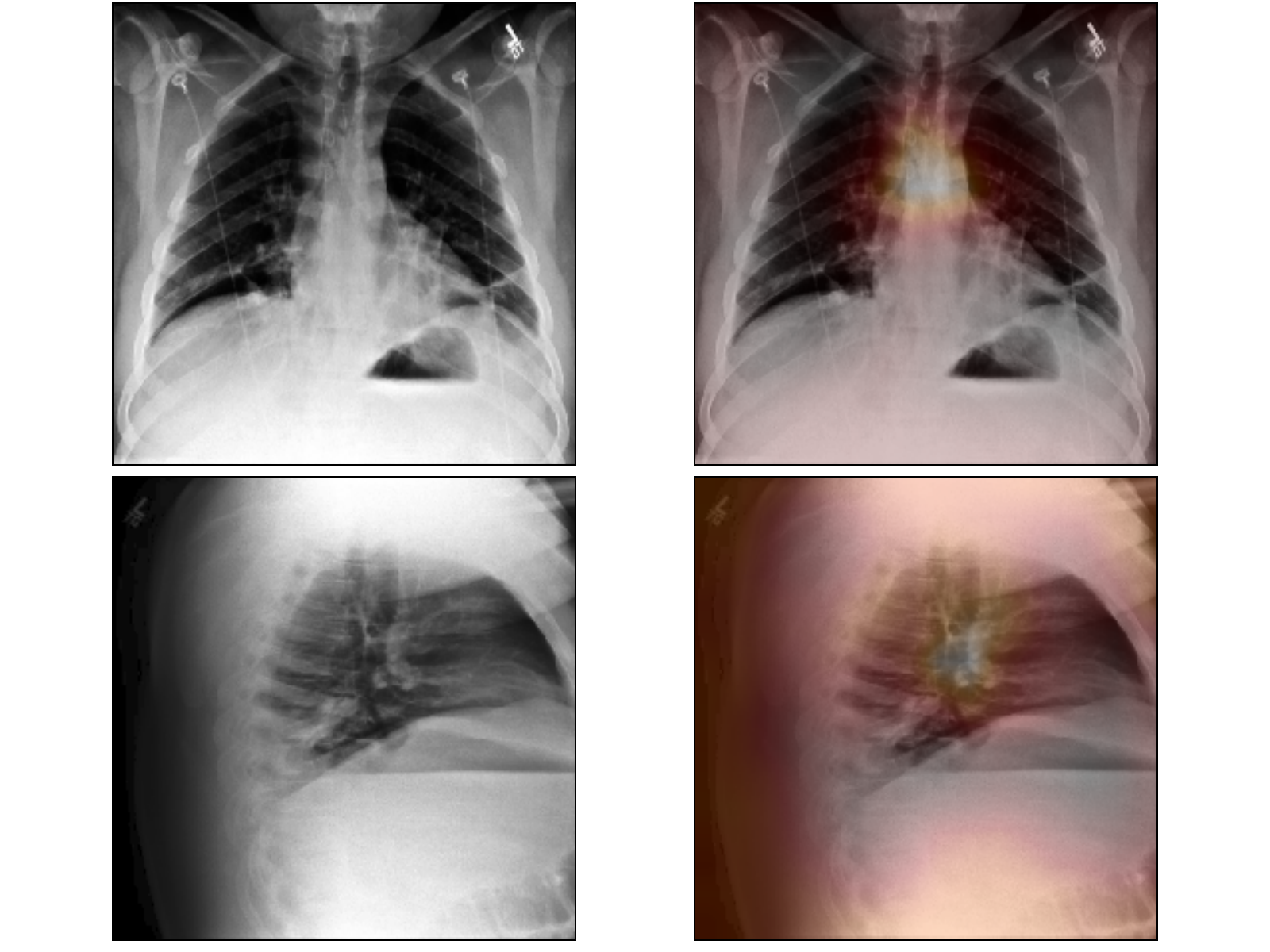}&
     \includegraphics[width=2in]{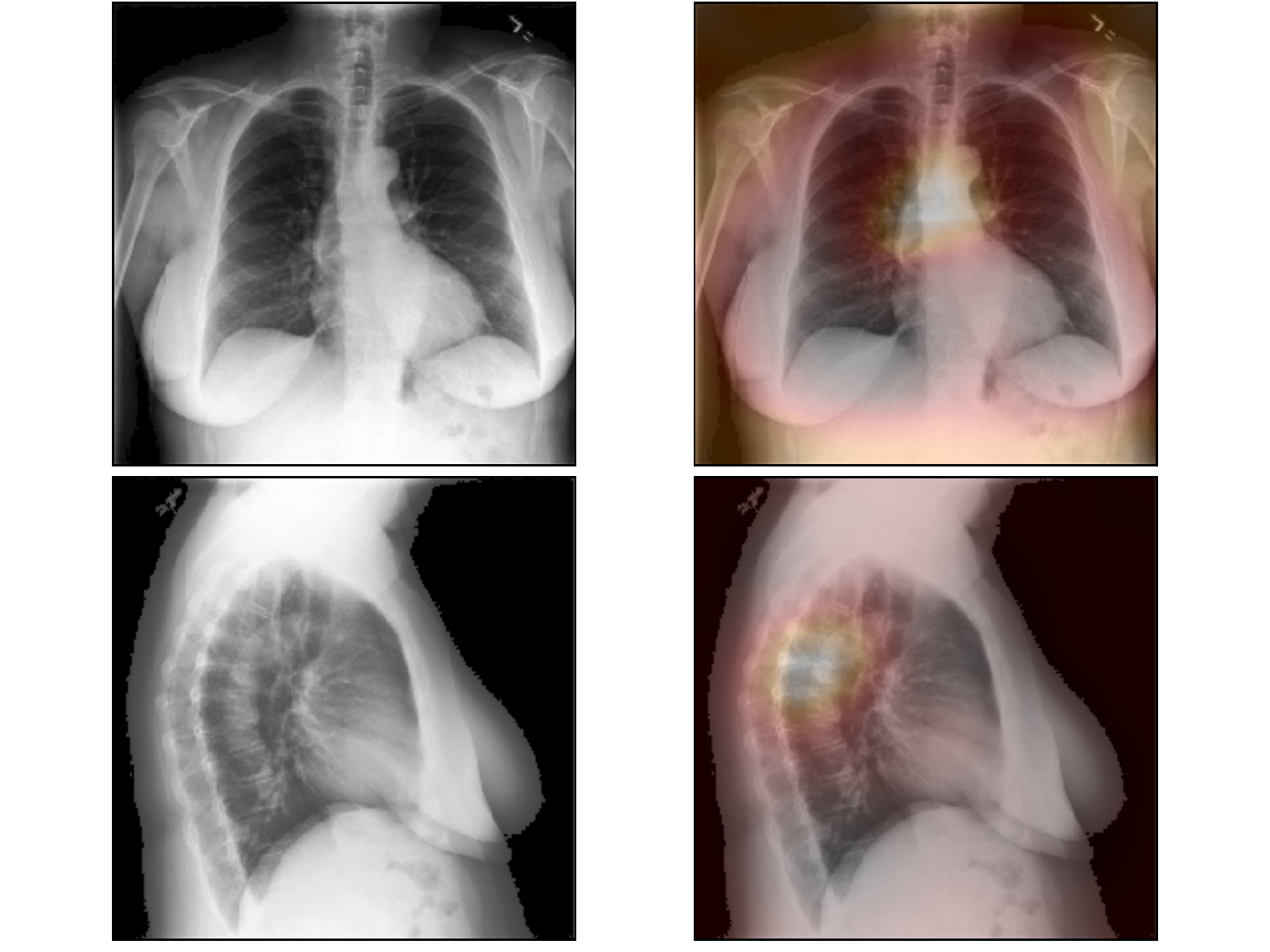}  \\\hline
        \quad\quad \textbf{GT}: no finding   & \quad\quad \textbf{GT}: pulmonary atelectasis, opacity&\quad\quad \textbf{GT}: scoliosis / degenerative\\
        \quad\quad \textbf{Pred}: no finding & \quad\quad \textbf{Pred}: hypoinflation & \quad\quad \textbf{Pred}: scoliosis / degenerative\\
        \hline
        \end{tabular}
    \caption{Visualization of Pathological Spatial Attention Map. \textbf{GT} represents the ground-truth disease labels of the image and \textbf{Pred} refers to the predicted disease categories}
    \label{fig:Pathological Spatial Attention}
\end{figure*}

\figureautorefname~\ref{fig:Pathological Spatial Attention} visualizes the attention score obtained by the pathological attention mechanism, i.e., Eq.~(3) as described in Section 3.1.2. We observed that pathological attention mechanism learns to accurately locate the disease visual attributes that associates with the predicted disease categories the in the original image. This experiment further demonstrates the pathological attention is effective in extracting disease related image features.
\begin{figure*}[htb]
    \centering
    \includegraphics[width=\linewidth]{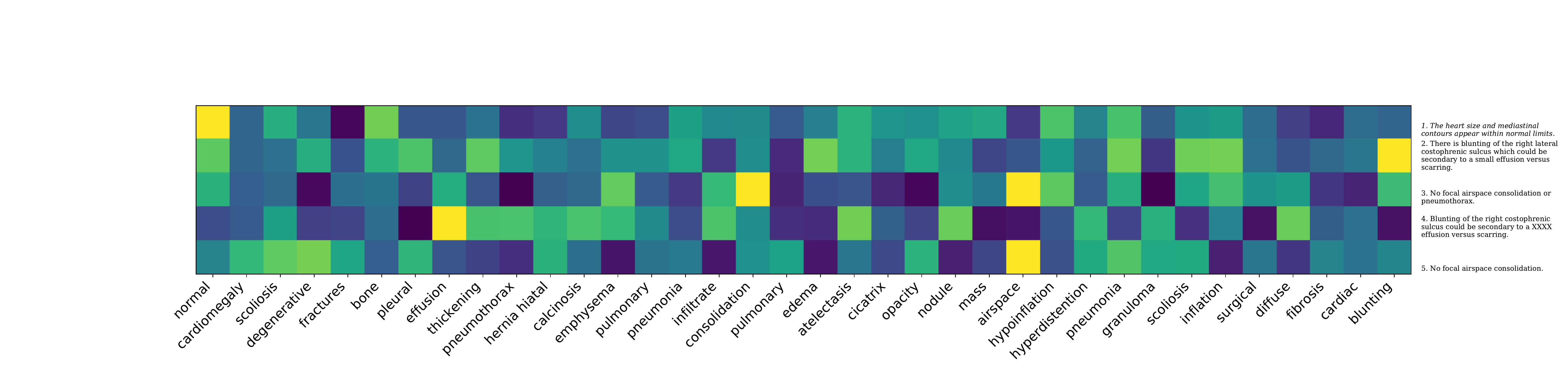}
    
    \includegraphics[width=\linewidth]{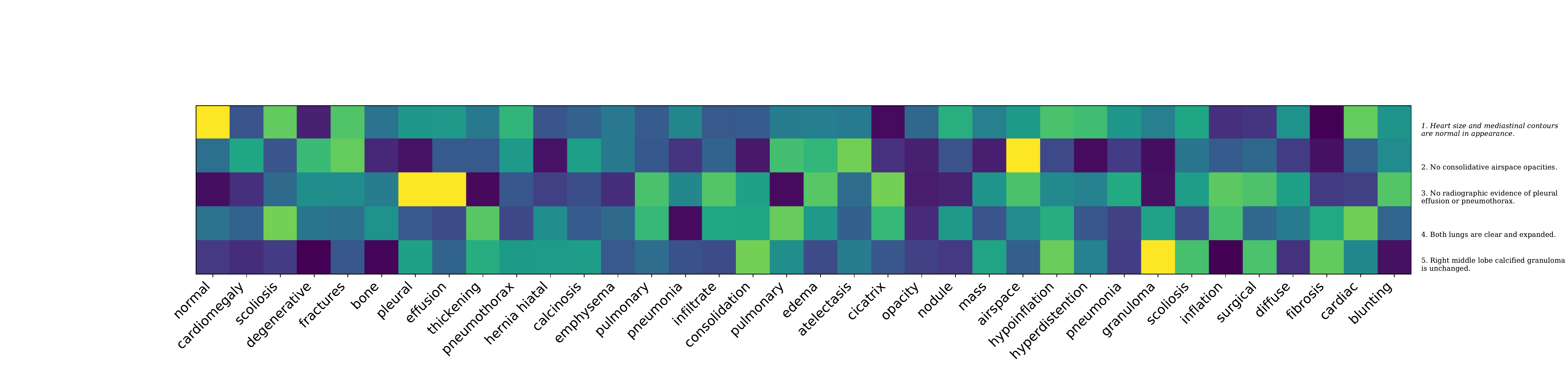}
    \caption{Visualization of multi-query attention score map between retrieved sentences and medical keywords}
    \label{fig:Multi-query attention}
\end{figure*}
\begin{figure*}
    \begin{center}
        \scriptsize
    \begin{tabular}{m{0.11\textwidth} m{0.12\textwidth} p{0.3\textwidth}p{0.35\textwidth}p{0.3\textwidth}}

       \textbf{Frontal Image} & \textbf{Lateral Image}  & \textbf{Ground Truth} & \textbf{Retrieved Report }  
\\\hline
         \includegraphics[width=0.8in,height=0.8in]{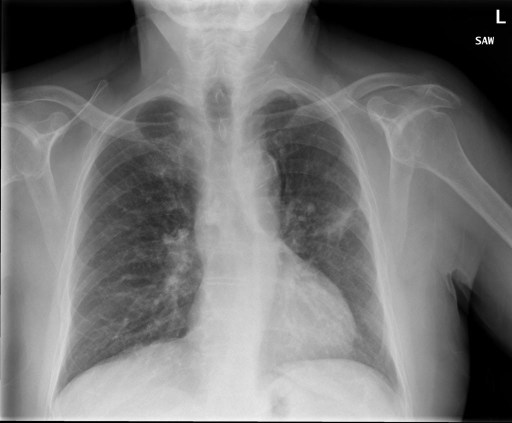} &  \includegraphics[width=0.8in,height=0.8in]{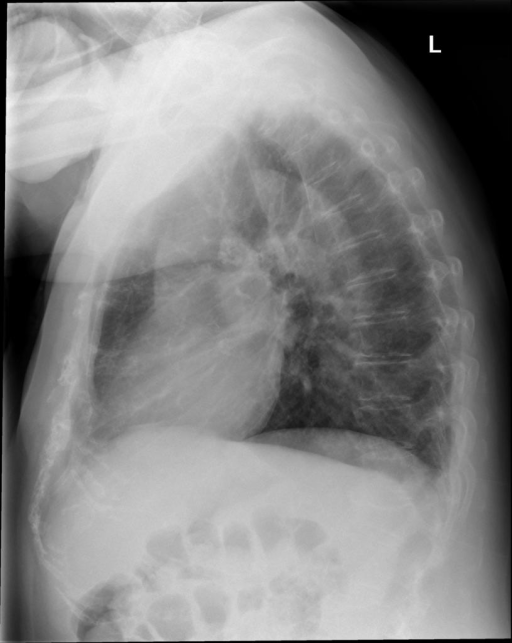} & \pbox{5cm}{\textcolor{red}{left midlung opacity} may be secondary to acute infectious process or developing mass lesion. followup to resolution is recommended. the heart is normal in size. the mediastinal contours are stable. aortic calcifications are noted. there are small calcified lymph $\langle\textsc{unk}\rangle$. emphysema and chronic changes are identified. there is $\langle\textsc{unk}\rangle$ opacity in the left perihilar upper lobe. \textcolor{red}{there is questionable $\langle\textsc{unk}\rangle$ $\langle\textsc{unk}\rangle$ to the pleural $\langle\textsc{unk}\rangle$}. this may represent acute infiltrate or developing density. there is no pleural effusion or pneumothorax. 
}
        &  \pbox{6cm}{cardiomegaly with central pulmonary vascular congestion. no xxxx edema. the heart is significantly enlarged. prominent pulmonary vascularity. no focal airspace consolidation, \textcolor{red}{suspicious pulmonary opacity}, or definite pleural effusion. no pneumothorax. visualized osseous structures appear intact.} 
\\\hline

         \includegraphics[width=0.8in,height=0.8in]{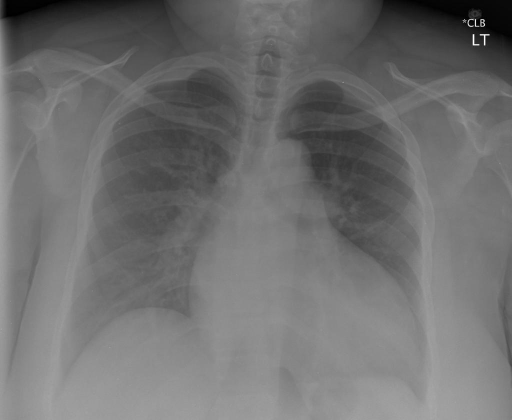} &  \includegraphics[width=0.8in,height=0.8in]{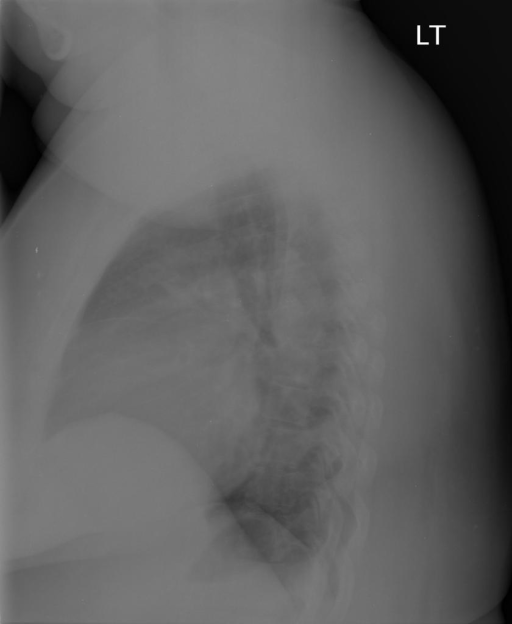} & \pbox{5cm}{\textcolor{red}{stable moderate cardiomegaly with prominent central pulmonary vasculature}. improved left basilar atelectasis or infiltrate. stable moderate cardiomegaly. mediastinal contours are unchanged. \textcolor{red}{stable prominence of the central pulmonary vasculature} with coarse central interstitial markings. decreased left basilar airspace disease. no visible pleural effusion or pneumothorax. 
}
        &  \pbox{6cm}{\textcolor{red}{cardiomegaly with central pulmonary vascular congestion.} no xxxx edema. the heart is significantly enlarged. \textcolor{red}{prominent pulmonary vascularity}. no focal airspace consolidation, suspicious pulmonary opacity, or definite pleural effusion. no pneumothorax. visualized osseous structures appear intact.
} 
\\\hline
    \end{tabular}
    \end{center}
    \caption{Examples of ground-truth report and retrieved reports by our VLR module. Highlighted \textcolor{red}{red} phrases are medical abnormality terms that generated and ground-truth reports have in common. \textbf{Bold} terms are common descriptions of normal tissues.}
    \label{fig:retrieved reports}
\end{figure*}
\subsection{Multi-query attention}
\figurename~\ref{fig:Multi-query attention} provides the visualization of the multi-query attention score map between medical keywords and retrieved template sentences. The brighter yellow represents the keyword-sentence pair has high attention score. We may make the following observation: The multi-query attention mechanism accurately highlight the sentences contains the keywords, e.g. ``normal'' in Case 1 sentence 1, ``airspace'' in Case 1 sentence 5, ``effusion'' and ``pleural'' in Case 2 sentence 3. This illustrates that the proposed multi-query attention is capable to highlight the key information in the medical description, thus assisting the feature extraction of the retrieved reports and sentences.

\subsection{Retrieved reports}
One of the advantages of the proposed {\name} is to automatically retrieve relevant medical reports from the training corpus based on the given image query. 
Examples of retrieved reports using the VLR module are shown in \figurename~\ref{fig:retrieved reports}. We only provide \textit{one retrieved report} that best correlates with each visual query for presentation clarification.

\subsection{Retrieved sentences}
The state-of-the-art retrieval-based approaches, e.g. HRGR-Agent~\cite{li2018hybrid} and KEPP~\cite{li2019knowledge}, 
generate standardized reports with \textit{a predefined retrieval database}, which is shared and fixed for all the input images. 
However, the proposed {\name} can \textbf{dynamically} learn such templates based on the data.
We provides some sentence-level retrieval results provided by the LLR module in Table~\ref{tab:Retrieved sentences}. For each query sentence, we show the 5 most relevant sentences in the retrieved reports using the VLR module. We can see that, the retrieved sentences provide key pathology or location information for the next sentence. 
This shows that the extracted relevant sentences provide useful additional information, which help {\name} to generate the next sentence of medical report.
\begin{table*}
    \centering
    \scriptsize
    \begin{tabular}{c|c|c}
    \hline
        \textbf{Query Sentence} & \textbf{Answer Sentences}& \textbf{Ground Truth Next Sentence} \\\hline
         \pbox{8cm}{There is haziness in the right lung apex.} & \pbox{4cm}{1. apparent \textcolor{red}{nodular opacity} on lateral projection, immediately retrocardiac, is xxxx to represent confluence of overlapping silhouettes. \\
         2. visualized osseous structures intact. \\
         3. prosthetic right humeral head. \\
         4. at the \textcolor{red}{right lung} apex, there is a more \textcolor{red}{focal ovoid lucency which measures approximately 1.3 cm}. \\5. lucent lesion with thin sclerotic margin in the right humeral head.}& \pbox{4cm}{There is a \textcolor{red}{1.7 cm nodular density in the medial right lung} base seen on the frontal view, not identified on the lateral view.}\\\hline
         Diffuse, right greater than left, interstitial opacities. & \pbox{4cm}{1. tortuous and dilated \textcolor{red}{aorta}. \\
         2. there is questionable dilation of the pulmonary \textcolor{red}{arteries}. \\
         3. there is increased kyphosis of the thoracic spine similar to the prior study \\
         4. large hiatal hernia with dilated intrathoracic stomach. \\5. abdomen: there are no dilated loops of bowel to suggest obstruction.}& \pbox{4cm}{Central \textcolor{red}{vascular} congestion.}
         \\
         \hline
    \end{tabular}
    \caption{Retrieved sentences using LLR module. The medical phrases marked by \textcolor{red}{red} correctly forecast the information in the next sentence.}
    \label{tab:Retrieved sentences}
\end{table*}

\subsection{Generated reports}
\begin{figure*}
    \begin{center}
    \scriptsize
    \begin{tabular}{m{0.11\textwidth} m{0.12\textwidth} p{0.22\textwidth}p{0.22\textwidth}p{0.2\textwidth}}

       \textbf{Frontal Image} & \textbf{Lateral Image}  & \textbf{Ground Truth} & \textbf{MvH+AttL}~\cite{yuan2019automatic} & \textbf{{\name}}   
\\\hline
         \includegraphics[width=0.8in,height=0.8in]{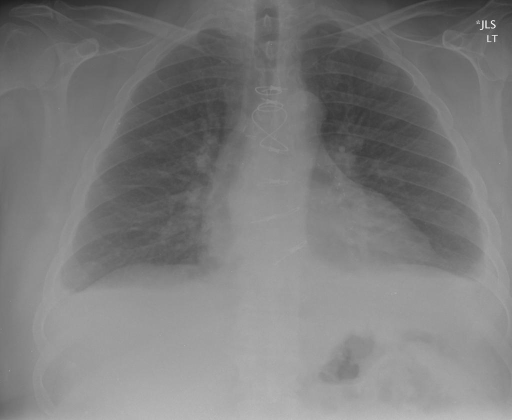} &  \includegraphics[width=0.8in,height=0.8in]{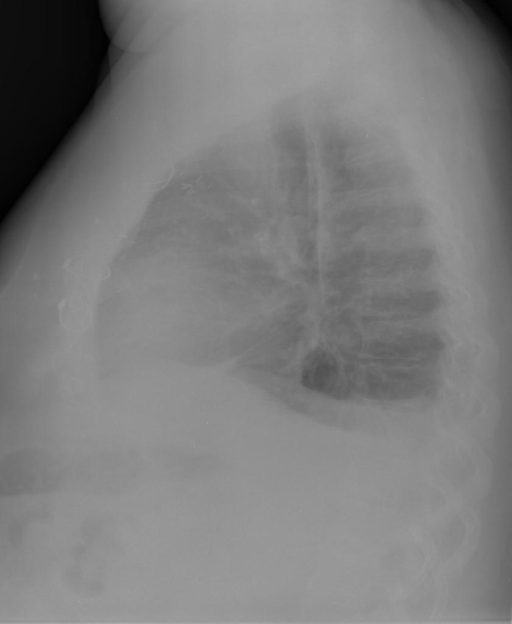} & \pbox{3.5cm}{\textcolor{red}{mild pulmonary vascular congestion} with $\langle\textsc{unk}\rangle$ $\langle\textsc{unk}\rangle$ bilateral effusions. $\langle\textsc{unk}\rangle$ findings is most compatible with congestive heart failure. pa and lateral views. stable postoperative changes with midline sternotomy $\langle\textsc{unk}\rangle$ and $\langle\textsc{unk}\rangle$ $\langle\textsc{unk}\rangle$. cardiac size remains mildly enlarged but stable. \textcolor{red}{there is mild vascular congestion}. small bilateral pleural effusions are present which are $\langle\textsc{unk}\rangle$. 
}
        &  \pbox{3.5cm}{cardiomegaly with mild bibasilar airspace disease. is clear. there is mild cardiomegaly. normal limits. effusion.   $\langle\textsc{unk}\rangle$ are grossly normal.
} 
        & \pbox{3.5cm}{low lung volumes with \textcolor{red}{bronchovascular crowding}. no acute infiltrate or atelectasis. there is an $\langle\textsc{unk}\rangle$ opacity in the left midlung which may represent early $\langle\textsc{unk}\rangle$ or $\langle\textsc{unk}\rangle$. no pneumothorax or $\langle\textsc{unk}\rangle$ pleural effusions. the cardiac and mediastinal contours are within normal limits.
}
\\\hline
         \includegraphics[width=0.8in,height=0.8in]{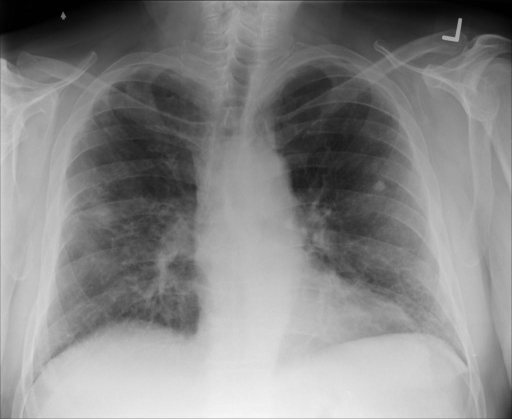} &  \includegraphics[width=0.8in,height=0.8in]{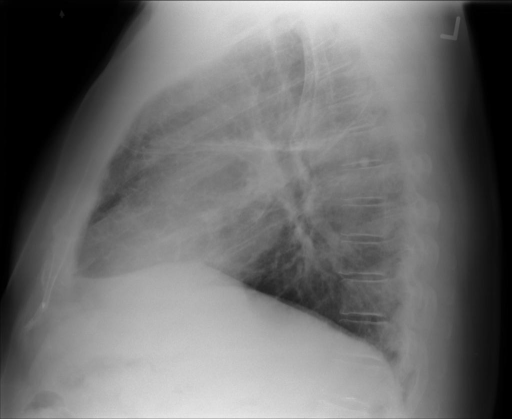} & \pbox{3.5cm}{round density within the anterior segment of the right upper lobe. this may represent $\langle\textsc{unk}\rangle$ pulmonary nodule. the primordial $\langle\textsc{unk}\rangle$ was $\langle\textsc{unk}\rangle$ to $\langle\textsc{unk}\rangle$ the $\langle\textsc{unk}\rangle$ $\langle\textsc{unk}\rangle$ of this critical finding. there is a calcified granuloma left midlung. there is round density within the anterior segment of the right upper lobe. there are \textcolor{red}{prominent interstitial opacities} which may represent changes associated with fibrosis. heart size is normal. no pneumothorax. anterior segment of upper lobe rounded focal density. could be $\langle\textsc{unk}\rangle$ lung nodule. 
}
        &  \pbox{3.5cm}{
    no acute cardiopulmonary abnormality. stable cardiomediastinal silhouette. there is mild tortuosity of the thoracic aorta. no focal airspace consolidation pneumothorax or pleural effusion. is noted in the right upper quadrant. $\langle\textsc{unk}\rangle$ are seen on the lateral view. bony abnormalities. 
} 
        & \pbox{3.5cm}{right upper lobe pneumonia. consideration and lateral views of the chest show normal size and configuration of the cardiac silhouette. normal mediastinal contour pulmonary $\langle\textsc{unk}\rangle$ and vasculature central airways and aeration of the lungs. no pleural effusion. there are $\langle\textsc{unk}\rangle$ \textcolor{red}{right $\langle\textsc{unk}\rangle$ $\langle\textsc{unk}\rangle$ streaky opacity}. there is no large pleural effusion. 
}
 \\\hline
        \includegraphics[width=0.8in,height=0.8in]{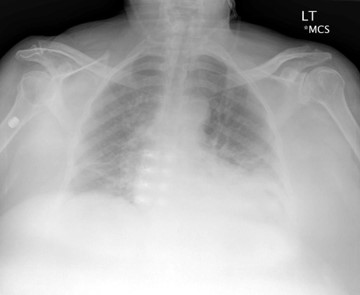} &  \includegraphics[width=0.8in,height=0.8in]{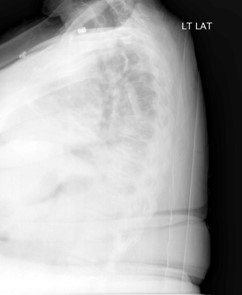} &\pbox{3.5cm}{left lower lobe pneumonia and minimal scarring or subsegmental atelectasis in the right lung base. \textcolor{red}{overall low lung lines.} there is scarring or subsegmental atelectasis at the right lung base. in the \textit{left lower lobe} there is \textcolor{red}{airspace disease} consistent with pneumonia. no pneumothorax. \textit{heart and mediastinum are stable given the lung volumes.} \textcolor{red}{degenerative changes in the spine.} } &\pbox{3.5cm}{\textit{cardiomegaly and pulmonary vascular congestion. right} \textcolor{red}{basilar airspace disease.} no definite pleural effusion. stable cardiomegaly. the thoracic aorta is tortuous. pulmonary vascularity is within normal limits. no pneumothorax or pleural effusion.  } & \pbox{3.5cm}{\textcolor{red}{low lung volumes otherwise clear}. stable cardiomediastinal silhouette. low lung volumes. without focal consolidation pneumothorax or pleural effusion. \textcolor{blue}{limited lateral view given overlapping tissue silhouettes.} \textcolor{red}{degenerative changes of the thoracic spine.} } \\\hline
       
    \end{tabular}
    \end{center}
    \caption{Examples of ground-truth report and generated reports by MvH+AttL~\cite{yuan2019automatic} and {\name}. Highlighted \textcolor{red}{red} phrases are medical abnormality terms that generated and ground-truth reports have in common. \textbf{Bold} terms are common descriptions of normal tissues. The text in \textit{italics} is the opposite meaning of the generated report and the actual report. We also mark the supplementary comments to the original report in \textcolor{blue}{blue}.}
    \label{fig:more example}
\end{figure*}

We visualize more examples for the generated reports on the Open-i dataset in \figureautorefname~\ref{fig:more example}. First, we can see that our {\name} generally produces longer text compared with the baseline MvH+AttL~\cite{yuan2019automatic}. 
This phenomenon further validates the superiority of our method in long text generation tasks. At the same time, we also find that {\name} embodies more accurate lesion information in the text, which indicates the value of the proposed method {\name} in medical diagnosis. 

\bibliographystyle{acl_natbib}
\bibliography{anthology,acl2021}
